\documentclass[letterpaper, 10 pt, conference]{ieeeconf}  % Comment this line out if you need a4paper
\usepackage{booktabs}
\usepackage{epsfig} % for postscript graphics files
\usepackage{mathptmx} % assumes new font selection scheme installed
\usepackage{amsmath} % assumes amsmath package installed
\usepackage{amssymb}  % assumes amsmath package installed

\usepackage{subcaption}
\usepackage[dvipsnames]{xcolor}
\usepackage{cite}
\usepackage{comment}

\IEEEoverridecommandlockouts                              % This command is only needed if 
\title{\LARGE \bf
Distributed Virtual Model Control for 
Scalable \\ Human-Robot Collaboration in Shared Workspace
}

\author{Yi Zhang, Omar Faris, Chapa Sirithunge, Kai-Fung Chu, Fumiya Iida and Fulvio Forni% <-this % stops a space
\thanks{Y. Zhang and O. Faris are supported by the Engineering and Physical Sciences Research Council [EP/S023917/1]. This project has received funding from the European Union’s Horizon 2020 research and innovation programme under the Marie Skłodowska-Curie grant agreement No 101034337. The authors are with the Department of Engineering, University of Cambridge, Cambridge CB2 1PZ, United Kingdom
        {\tt\small \{yz892, of292, csh66, kfc35, fi224\}@cam.ac.uk, f.forni@eng.cam.ac.uk}}% <-this % stops a space
}

\begin{document}

\maketitle
\thispagestyle{empty}
\pagestyle{empty}

%%%%%%%%%%%%%%%%%%%%%%%%%%%%%%%%%%%%%%%%%%%%%%%%%%%%%%%%%%%%%%%%%%%%%%%%%%%%%%%%
\begin{abstract}

We present a decentralized, agent agnostic, and safety-aware control framework for human–robot collaboration based on Virtual Model Control (VMC). In our approach, both humans and robots are embedded in the same virtual-component-shaped workspace, where motion is the result of the interaction with virtual springs and dampers rather than explicit trajectory planning. A decentralized, force-based stall detector identifies deadlocks, which are resolved through negotiation. This reduces the probability of robots getting stuck in the block placement task from up to 61.2\% to zero in our experiments. The framework scales without structural changes thanks to the distributed implementation: in experiments we demonstrate safe collaboration with up to two robots and two humans, and in simulation up to four robots, maintaining inter-agent separation at around 20 cm. Results show that the method shapes robot behavior intuitively by adjusting control parameters and achieves deadlock-free operation across team sizes in all tested scenarios.

\end{abstract}

%%%%%%%%%%%%%%%%%%%%%%%%%%%%%%%%%%%%%%%%%%%%%%%%%%%%%%%%%%%%%%%%%%%%%%%%%%%%%%%%
\section{Introduction}

Human-robot collaboration (HRC) aims to bridge the gap between human dexterity and robot precision. Compared to multi-robot collaboration, HRC in shared workspaces must account for (i) unpredictability and complexity of human motion and (ii) life-critical cost of failure and injury \cite{2013Aude}. Hence, HRC literature mainly focuses on ensuring human safety via rich sensing \cite{saleem2025review}, human intent prediction \cite{hoffman2024inferring}, and optimization and learning-based algorithms \cite{li2024SafeHumanRobot, mukherjee2022SurveyRobotLearning, fan2024state}. 
Nonetheless, these methods can be computationally expensive, model- or data-dependent, and tailored mainly to humans.

In shared workspaces, agents should be able to enter and leave freely, sharing workloads and roles.
However, current approaches often treat humans and robots differently, applying unique rules for each, with robots typically adapting to human actions \cite{sanfilippo2025caged, mingyue2017human}. In this paper, we challenge this distinction. Our key question is: can we move towards a lightweight, agent-agnostic, and scalable human-robot collaboration while guaranteeing human safety?

We start from the idea that robot and human agents can have shared or complementary collaborative roles. We search for a decentralized control architecture that can shape the collective behavior without relying on accurate models or large datasets.
The architecture should have a low computational load and seamlessly integrate new agents without the need to redesign the control strategy. 
The goal is to derive a strategy that allows for any number of agents to participate in the task, or withdraw from it, as required, while guaranteeing a safe multi-agent collaboration of human or robots. 

For this purpose, we propose a safety-aware control framework based on Virtual Model Control (VMC).
In our approach, both humans and robots are embedded in a common workspace shaped with virtual components. 
Robot motion is regulated by virtual springs and dampers, without explicit trajectory planning. 
By adjusting virtual component parameters, we can vary the interaction behavior from fast to slow and change how cautious of other agents.
Human’s behavior is then influenced through their natural adaptation to the workspace. 
With nonlinear virtual springs, we bound the force applied by the robot upon potential physical interaction.

\begin{figure}[t]
  \centering
  \includegraphics[width=0.9\linewidth]{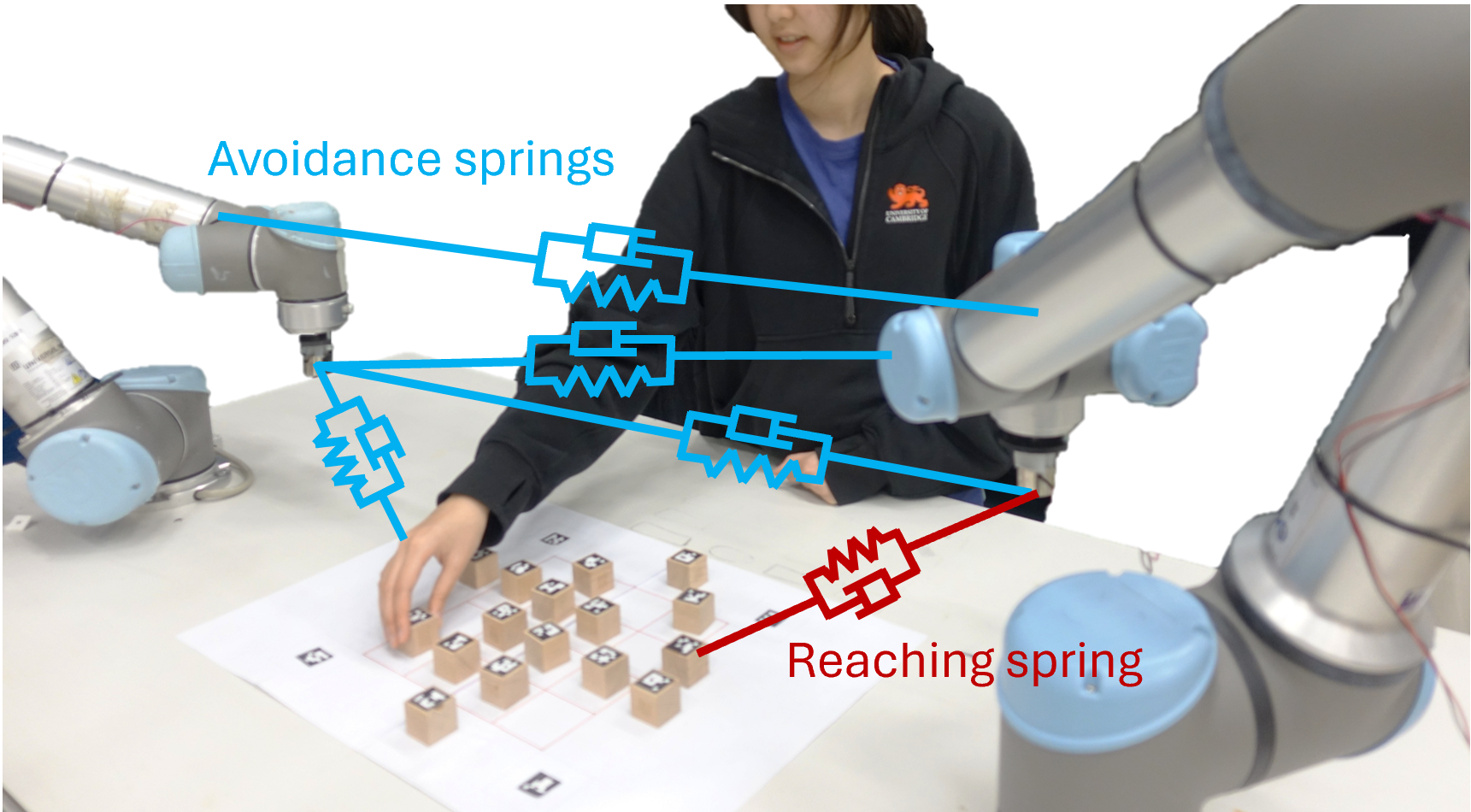}
  \caption{Human-robot collaboration in a pick and place task.}
  \vspace{-4mm}
  \label{fig:1}
\end{figure}

In this paper, we validate our approach in collaborative pick-and-place task involving both humans and robots, as seen in Fig. \ref{fig:1}.
Our contributions include
\begin{itemize}
\item An \textbf{agent-agnostic approach to multi-agent collaboration} where robots and humans are treated on the same ground. The robot behavior and its interaction with other robots and human are shaped by intuitive mechanical parameters. A separation distance of about 20 cm is mantained through active avoidance and compliance.
\item A \textbf{VMC-based implementation with conflict resolution}, capable of detecting deadlocks via force balance and negotiating priority through minimal communication, eliminates the failure mode of robots getting stuck during the block placement task, reducing its occurrence from a maximum of 61.2\% to zero experimentally.
\item \textbf{Scalability:} the decentralized approach allows seamlessly swapping of agents and smooth adaptation to changes in team composition, shown for up to two robots and two humans in experiment, and up to four robots in simulation. 
\end{itemize}

\section{Related Works}

\subsection{Virtual Model Control}

VMC offers an intuitive control framework where virtual mechanical components (e.g., springs and dampers) are used to design physics-driven predictable robotic motion. The relevance of VMC for this paper is 
that it moves beyond pre-programmed or planned motion. Instead, robot behavior is a direct consequence of its interaction with virtual mechanical components. This enables a novel design approach that is particularly well-suited for interactive tasks involving a variety of low-predictability scenarios.
VMC was originally introduced for bipedal locomotion in \cite{Pratt2001VirtualMC} and later adopted for quadruped locomotion \cite{chen2020virtual, winkler2014path}. Recently, VMC was utilized for robot manipulation scenarios such as reaching under uncertainties \cite{2024Zhang}, robot assisted laparoscopic surgery tasks \cite{larby_optimal_2025}, and robotic cutting \cite{zhang2025rock}.

In HRC, VMC was used in collaborative surgical bone drilling task for increased drill alignment accuracy \cite{larby2025collaborative}. In this work, we expand the use VMC towards multi-agent scenarios involving humans and robots. Additionally, we present the use of VMC in position‑controlled robotic manipulators, which have not been addressed before. With VMC, all agents can be abstracted as virtual mechanisms, resulting in a decentralized control architecture that enables swift addition or removal of any agent. Agents avoid each other with mechanisms like virtual repulsive springs and dampers. No path planning is required as trajectories emerge from these interactions. This also enables sparse communication among the robots, which is only truly required to handle stalls, detected through force-based analysis.

\subsection{Human–Robot Collaboration}

Safety in the shared workspaces is commonly addressed through motion planning guided by sensing and human intention prediction \cite{li2024SafeHumanRobot}, 
where robots slow down, re-route, or completely stop based on the human state.
Researchers in \cite{lyu2023EfficientCollisionFreeHuman} predicted human arm pose to schedule robot goals with a learning-based method in a collaborative pick-and-place task. 
Compliance with ISO/TS 15066 was addressed in \cite{pupa2024DynamicPlannerSafe} with an online optimizer that generated collision-free trajectories.
Potential fields with a reference governor were utilized for real-time safety in \cite{merckaert2024RealtimeConstraintbasedPlanning}.
Proximity sensing on the robot body is utilized for contact anticipation and avoidance in \cite{9636130}.
Fast distance evaluation is investigated in \cite{6225245} for avoiding moving obstacles.
Nonetheless, in close proximity during HRC, collision remains possible, and stiff robot behaviors can cause injury in such cases.

Impedance control \cite{hogan1984impedance} enables compliant interaction in uncertain environments and has been applied in shared workspaces to mitigate impact forces.
It has been combined with proximity sensing for safe, collision-free close interactions \cite{bertoni2025ProximityBasedFrameworkHumanRobot}, with potential fields for obstacle and human avoidance in assembly tasks \cite{ktistakis2025robot}, and with human-in-the-loop adaptation using stress and attention signals to generate personalized trajectories \cite{lagomarsino2025PROMINDProximityReactivity}.

The VMC of this paper and impedance control are similar in that both methods impose a desired dynamic behavior, like that of a spring-damper system.
The main difference is that VMC unifies motion planning and control by defining the robot's end-effector trajectory through its interaction with a set of user-designed virtual components. These virtual components (like springs and dampers) pull the robot towards its goal and push it away from obstacles. The final trajectory emerges from these interactions, eliminating the need for separate path-planning computation \cite{larby_optimal_2025, 2024Zhang}. In contrast, traditional impedance control is primarily a motion control method. It assumes a pre-planned trajectory and makes the robot's end-effector behave compliantly along that path.

\subsection{Multi-agent Collaboration}

Multi‑robot and human-robot collaboration both require the robot to avoid moving obstacles to manage physical contact. Previous works addressed collision avoidance between multiple robots in a shared workspace through motion coordination with trajectory replanning \cite{behrens2020simultaneous, zhang2023online} or reacting to repulsive actions from collision prevention regions, such as control barrier functions (CBFs), solved via quadratic programming \cite{sun2024real, CBF2019distributed, CBF2020, CBF2022distributed}. Unlike the method proposed here, these solutions are often centralized or, even if implemented distributively, rely on the global exchange of information and variables among all agents.

Human collaboration with multi-robot systems have been addressed before  \cite{dehio2022EnablingImpedancebasedPhysical, wen2023collaborative}. These works focused on physical collaboration rather than the safety of humans in shared workspaces and utilized centralized architectures. This motivated \cite{shi2024safe, shi2025distributed} to develop a decentralized CBFs approach for safe human interactions with multiple robots in the same workspace. However, the method formulation only considered collision avoidance with the human without accounting for potential inter-robot collisions. Additionally, as a CBFs-based approach, it relied on availability of information from all agents. There remains a gap for treating all collaborative agents on the same level with minimal information exchange, enabling seamless interactions among all of them.

\section{Methodology}\label{methodology}
\subsection{Virtual Mechanism for Reach and Avoidance}\label{sec:vm}

The robot's reach and avoidance motions are generated by the interaction between its dynamics and a virtual mechanism. This mechanism consists of a goal spring that pulls the end-effector toward the next task waypoint, avoidance springs that repel other agents and obstacles, and associated dampers.

A spring and a damper attached between end effector at $\pmb{x}_{EE}$ and goal at $\pmb{x}_{goal}$ produces the force:
\begin{align}
	\pmb{f}_{goal}(t) &= k\pmb{x}(t) + c\dot{\pmb{x}}(t), \quad \pmb{x}(t) = \pmb{x}_{goal}-\pmb{x}_{EE}(t) 
	\label{eq:goal_spring}
\end{align}
at each time $t$, where $k$ and $c$ are the stiffness and damping coefficients.
Gentle long-distance movements are obtained through filtering.
Given a strictly monotone time law $s:[0,T]\subseteq \mathbb{R} \to [0,1]$ such that $s(0)=0$ and $s(T)=1$, we take
\begin{align}
    \pmb{x}(t) &= \pmb{x}_{EE}(0) + s(t)(\pmb{x}_{goal}-\pmb{x}_{EE}(0))-\pmb{x}_{EE}(t), \; t\in [0, T]
\end{align}

\begin{figure}[ht]
    \centering
    \vspace{-2mm}
    \includegraphics[width=0.9\linewidth]{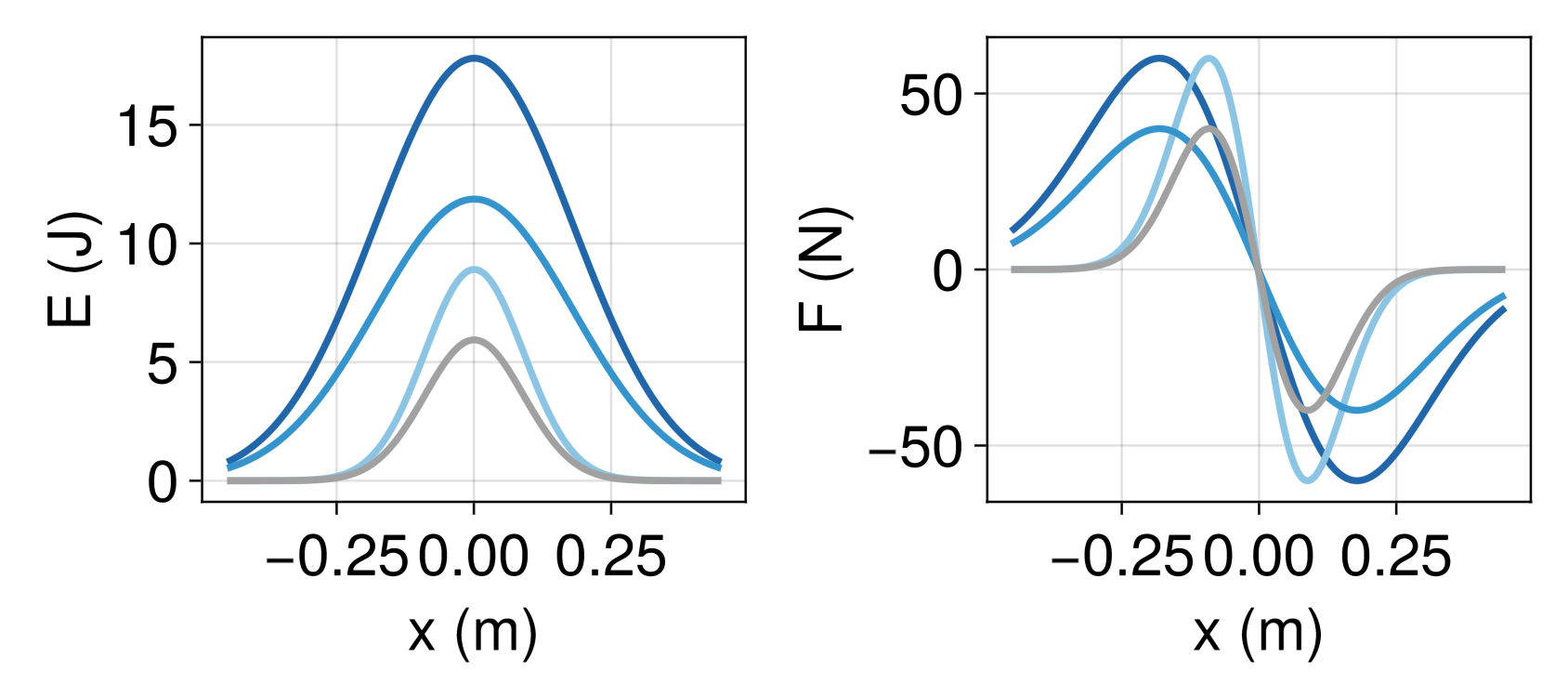}
    \vspace{-2mm}
    \caption{Four avoidance spring profiles. Profile 1 (gray): $f_{max}=-40, \, \sigma=0.09$; Profile 2 (light blue): $f_{max}=-60, \, \sigma=0.09$; Profile 3 (medium blue): $f_{max}=-40, \, \sigma=0.18$; Profile 4 (dark blue): $f_{max}=-60, \, \sigma=0.18$.}
    \label{fig:avoidance_spring_profile_4}
     \vspace{-1mm}
\end{figure}

For collision avoidance, we use springs with Gaussian energy profile $E(\pmb{x})=-k\sigma^{2}\exp\Bigl(-\frac{\lVert\pmb{x}\rVert^{2}}{2\sigma^{2}}\Bigr)$
 which generate local repulsive force
\begin{align}
    \pmb{f}_{avoid}(t) &= k\pmb{x}\!(t)\!\exp\!\Bigl(\!-\frac{\!\lVert\!\pmb{x}\!(t)\!\rVert^{2}}{\!2\sigma^{2}}\!\Bigr)\! ,\; \pmb{x}(t) = \pmb{x}_{obj}(t)-\pmb{x}_{EE}(t)
    % , \sigma = f_{max}e^{0.5}/k
    \label{eq:gaussian_energy}
\end{align}
% \end{subequations}
Here, denoting by $f_{max}$ the maximum force parameter, $\sigma=f_{max}e^{0.5}/k$ is the distance from the 
obstacle $\pmb{x}_{obj}$ at which this maximum force occurs.
Any two of $k, f_{max}, \sigma$ uniquely determine the spring profile.
Four examples of avoidance spring profile with varying $f_{max}$ and $\sigma$ are illustrated in Fig. \ref{fig:avoidance_spring_profile_4}.
To implement inter-robot avoidance, twelve virtual points are distributed over each robot body, and avoidance springs are attached between every pair of points belonging to different robots.
Additional avoidance springs are attached between the robot and other static obstacles, such as the table, with increased effective height when a block is grasped by the end-effector.

For human hand avoidance, we also take advantage of unilateral saturating dampers.
This element guarantees that a repulsive force is applied to the end effector (only) when a human hand is both (i) moving towards the end effector and (ii) within a specified proximity.
Unlike potential field and control barrier functions which continuously enforce a minimum separation, the unilateral damper penalizes relative velocity. Any motion of the robot and hand towards each other is actively damped, allowing for proximity but only at low relative speed. If we denote the hand position by $\pmb{x}_{hand}$, we have 
\begin{subequations} 
\begin{equation}
    \dot{r}(t) = \frac{\pmb{d}(t)^{\mathsf{T}}}{\|\pmb{d}(t)\|} \dot{\pmb{d}}(t) \quad \text{where} \quad\pmb{d}(t) = \pmb{x}_{hand}(t) - \pmb{x}_{EE}(t).
\end{equation}
The damper is active only when \(\dot{r} < 0\) (approaching motion); otherwise no force is applied. The force acting on the end effector is then given by
\begin{equation}
\pmb{f}_{brake}(t) =
\begin{cases}
\min\!\big(-c(\lVert\pmb{d}(t)\lVert)\dot{r}(t),\;f_{\max}\big)\,\frac{\pmb{d}(t)}{\|\pmb{d}(t)\|}, & \dot{r} < 0,  \\[4pt]
0, & \dot{r} \ge 0,\\
\end{cases}
\end{equation}
where $f_{\max}$ is the maximum allowable magnitude of the avoidance force and the damping coefficient $c$ is given by
\begin{equation}
    % c(\|\pmb{d}\|) = c
    c(\|\pmb{d}(t)\|) =
    \begin{cases}
    c_0\left(1 - \dfrac{\|\pmb{d}(t)\|}{R}\right), & \|\pmb{d}(t)\| < R, \\[8pt]
    0, & \|\pmb{d}(t)\| \ge R,
    \end{cases}
\end{equation}
\end{subequations} 
The damper is only activated when the hand is within the activation radius $R$ of the robot, and the damping coefficient $c$ decays linearly with the distance $\|\pmb{d}\|$ from its maximum value $c_0$ at contact to zero at $R$, ensuring a smooth transition.

To implement our virtual component-based controller, we used the VMRobotControl.jl package \cite{VMRobotControl}, a platform-independent framework designed for virtual model control in robotic systems. 
This package enables the construction and simulation of both the robot and virtual components such as masses, springs, and dampers. 
Given a robot's URDF file and virtual components' definition, the package computes the robot and virtual mechanism's dynamics.

\subsection{Stall Detection and Conflict Resolution} \label{stuck_detection}
Robot may face deadlock when the ensemble of forces acting on the robot results in a small equivalent force. This is a sign of conflicting requirements. To detect this deadlock, we consider the following metric
\begin{align}
\rho &= \lVert F_{\text{net}} - F_{\text{tot}} \rVert ,\qquad
F_{\text{net}} = \left\lVert \sum_{i=1}^{N} \pmb{f}_{i}\right\rVert, \,\, F_{\text{tot}} = \sum_{i=1}^{N} \lVert \pmb{f}_{i} \rVert,
\label{eq:rho}
\end{align}
where $\pmb{f}_{i}$ denotes the force generated by the $i$-th virtual component (goal or avoidance), and $N$ is the number of components attached to the robot. 
When $\rho$ exceeds a predefined threshold, the robot is 
considered in \emph{stall} state, which leads to a negotiation procedure.

All robots in stall undergo a prioritization procedure, whose goal is to resolve the deadlock. 
The robot to prioritize is selected according to the following rules, which are considered only if no robot is currently prioritized.
(i) Robots that have finished all tasks and are in rest position will not be prioritized.
(ii) Robots executing a grasping or releasing motion are prioritized if they are within a proximity threshold of the closest robot. 
(iii) Robots that are very close to their target (within 6 cm) are given priority.
If multiple robots satisfy (ii) or (iii), one is selected according to rule (iv).
(iv) For all remaining robots, priority is assigned via a biased draw to guarantee fairness.
Specifically, when more than two robots are stalled, one robot is prioritized according to the probability distribution:
\begin{equation}
\mathbb{P}(i) \;=\; 
\frac{\exp\left(-\alpha c_i\right)}{\displaystyle\sum_{j=1}^{n_{\text{robot}}}\exp\left(-\alpha c_j\right)}, 
\quad \alpha>0,
\end{equation}
where $c_i \in \mathbb{Z}$ is the number of times robot $i$ has previously been prioritized and $\alpha$ is a parameter that controls how strongly past prioritizations reduce its future likelihood of selection.
Once a robot is selected, its prioritization counter $c_i$ is updated, and the procedure repeats next time robots are stalled. Biased draw in (iv) ensures fairness: robots with fewer prior selections are more likely to be chosen, balancing workload over time.

During prioritization, the selected robot proceeds directly toward its goal without avoiding other robots (still avoiding humans), while all other robots suspend goal-directed motion and only perform avoidance. After the prioritized robot reaches its goal, all robots resume their original behaviors. 
In our experimental setting, all robots participate in prioritization since they operate in close proximity. In larger environments, however, prioritization can be restricted to local robots within a certain neighborhood. Note that (ii) is also applied when robots are not stalled to ensure safe and reliable block manipulation.
While rare edge cases might still exist in coupling VMC with our prioritization procedure, deadlocks are always resolved in our experiments. A formal analysis is left to an extended journal version of this paper.

Conflict resolution is applied only to robots. Humans are naturally capable of assessing the situation and deciding an appropriate time to engage with the workspace, and therefore do not encounter deadlocks when collaborating with others. In contrast, robots with only the mechanical layer (VMC) lack such ability, they would remain stalled in deadlock situations. The conflict resolution layer thus provides robots with minimal decision-making to either take the lead or compromise. Human safety is always preserved by maintaining robot–human avoidance regardless of prioritization.

\section{Implementation and experimental platform}\label{implementation}
Two 6-DoF manipulators (UR5) are positioned on opposite sides of a table containing a $4\times4$ grid ($24 \,\text{cm} \times 24 \,\text{cm}$). Sixteen cubic blocks ($3 \,\text{cm}$) are distributed around the grid, as illustrated in Fig. \ref{fig:1}. The task is to place all blocks into the grid cells. Placement proceeds from the center outward: robots first fill cells closer to the grid center; among cells at equal distance, selection is random. This ordering mimics an assembly process where central positions are completed before peripheral ones. Each robot initially selects blocks on its own side of the workspace and then proceeds to pick from the remaining blocks.
For each pick-and-place action, task waypoints ($\pmb{x}_{goal}$) are the block’s grasping position and the position above it.

An Intel RealSense D405 RGB-D camera tracks 21 human hand keypoints using MediaPipe \cite{zhang2020mediapipe} at 10 Hz for human avoidance. 
Avoidance performance of the robot is limited by the update rate of hand detection.
Control and perception run on a standard PC.

The implementation of VMC on UR5s is based on a simulated virtual UR5 robots, using the VMRobotControl.jl package \cite{VMRobotControl}. The interaction forces outlined in Section \ref{sec:vm} drive the virtual UR5 robots. The simulated physics determines the motion and provide a reference configuration to each real UR5.
Virtual springs and dampers for reaching and avoidance are connected (and/or disconnected during prioritization) between each virtual robot and relevant entities, such as task targets, physical robot's position and the workspace boundaries.
At each control cycle, we simulate the virtual robot’s dynamics under all virtual forces for a duration of $\Delta t$ to obtain a joint‑space target $\pmb{q}_d$.
This target is then sent to the physical robot.
Stiffness and damping gains are selected experimentally to ensure sufficient repulsive force for reliable obstacle and human avoidance.

\begin{figure}[htbp]
    \centering
    \vspace{-4mm}
    \includegraphics[width=0.6\linewidth]{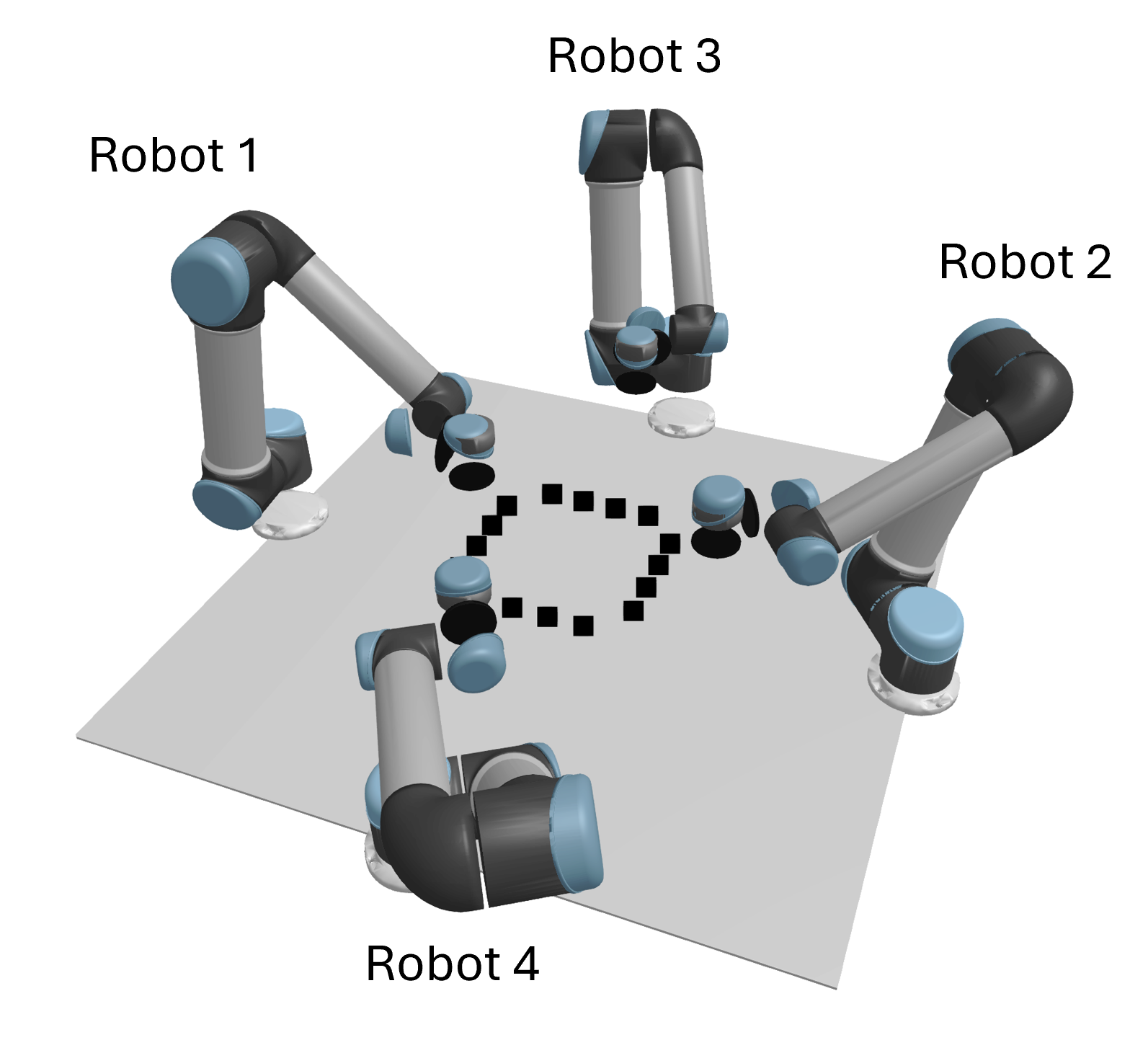}
    \caption{Multi-robot simulation. Robot 1 and 2 are for two-robot experiment. Robot 1-3 are for three-robot experiment.}
    \label{fig:four_robot_setup_sim}
    \vspace{-3mm}
\end{figure}

To assess scalability, we also conducted simulation tests with up to four robots. The setup is shown in Fig. \ref{fig:four_robot_setup_sim}.

\section{Experiment and results}\label{experiments}

\subsection{Agent-agnostic Collaboration}\label{sec:exp1}

We demonstrate the collaboration between a robot and either another robot or a human.
% using shared motion primitives (virtual springs).
In this experiment, Robot 1 tries to maintain its position in the center of the grid, while the other agent (robot or human) places all blocks.
Robot reaching is implemented through attractive springs.
Robot 1 avoids Robot 2 or the human via avoidance springs. 
In contrast, Robot 2 adopts an aggressive behavior and does not avoid Robot 1.
No communication or any form of negotiation is implemented between the agents. Robot 1 only perceives the position of human hand and the other robot. The parameters of the virtual mechanisms are in Table \ref{tab:control_parameters}.

\begin{figure}[ht]
    \centering
    \includegraphics[width=0.41\linewidth]{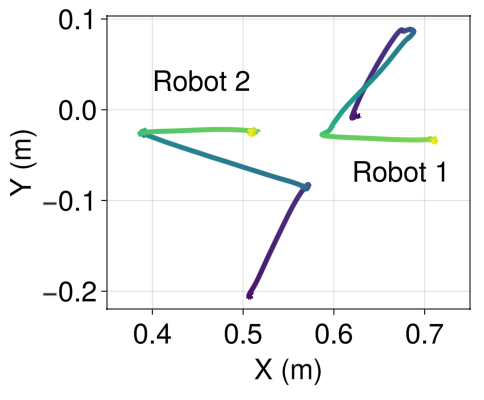}    
    \includegraphics[width=0.56\linewidth]{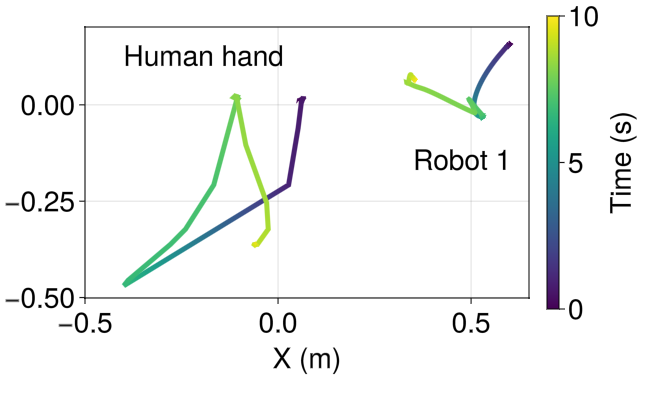} \vspace{-2mm}
    \caption{
    Robot 1 avoiding Robot 2 (left) and a human (right) with the same VMC avoidance spring. In both cases, Robot 1 retreats as the other agent approaches, then returns.
    }
    \vspace{-3mm}
    \label{fig:avoidance}
\end{figure}

Fig. \ref{fig:avoidance} presents sample trajectories from these experiments. Color indicates time, with matching colors 
identifying synchronous events. No deadlock occurred. Potential conflicts were resolved at the mechanical level, taking advantage of the different characteristics of each robot, due to the tuning of their control parameters.
Fig. \ref{fig:avoidance} shows that Robot 1 successfully avoided both Robot 2 and human hand while using the same virtual mechanism, demonstrating agent-agnostic behavior. 

To quantify spatial clearance, we define the minimum robot–robot distance $d_{\min}^{RR}$ as the smallest Euclidean distance between the end-effectors of two robots, and the minimum robot–human distance $d_{\min}^{RH}$ as the smallest distance between a robot end-effector and the closest tracked point of the human hand.
The minimum inter-robot distance was $d_{\min}^{RR}=16.6\,\text{cm}$, and the minimum robot–human distance averaged over five trials with different participants was $d_{\min}^{RH}=21.8\,\text{cm}$, confirming that safe separation was consistently maintained during interaction.

\subsection{Safety-aware Collaboration}\label{sec:exp1-2}
We investigate how different spring profiles influence the robot behavior.
The task is the same as the last section, within the setting of Fig. \ref{fig:1}.
Four avoidance spring profiles are illustrated in Fig. \ref{fig:avoidance_spring_profile_4}. 
Increasing $\sigma$ extends the range of repulsive force, while a larger $f_{\max}$ increases the force magnitude. 
We ran five trials with different participants for each spring profile.
Participants rested and re-entered the scene at least four times to encourage interaction with the robot.
The resulting minimum distances $d_{\min}^{RH}$ are in Table
\ref{tab:avoidance_metrics}.

We also examined Speed and Separation Monitoring (SSM) requirements, as specified in ISO/TS 15066 \cite{ISO15066}.
To illustrate how parameter tuning can be used to meet SSM requirements, we computed a conservative protective separation distance $S_{p}$
given by
\begin{align}
S_{p}=S_{h}+S_{r}+S_{s}+C+Z_{d}+Z_{r},
\end{align}
where
$S_{h}=v_{h}(T_{r}+T_{s})$ accounts for human motion,
$S_{r}=v_{r}T_{r}$ for robot reaction time,
$S_{s}$ for robot stopping distance,
$C$ for intrusion distance,
and $Z_{d}$ and $Z_{r}$ for uncertainty margins on human sensing and robot pose.  
Considering the size of the working space, we assumed max human speed in the direction of robot $v_{h}=0.8\;\mathrm{m\,s^{-1}}$,
max robot speed in the direction of operator $v_{r}=0.3\;\mathrm{m\,s^{-1}}$, 
controller reaction $T_{r}=0.08\;\mathrm{s}$, 
mid-speed stopping time $T_{s}=0.15\;\mathrm{s}$ (UR5 datasheet), 
$C=0.08\;\mathrm{m}$ for hand-resolution sensing, 
and sensor / pose uncertainties $Z_{d}=Z_{r}=0.02\;\mathrm{m}$.  
Taking $S_{s}\simeq\frac12 v_{r}T_{s}$ yields
$S_{p} \simeq 0.35\ \text{m}$.
Table \ref{tab:avoidance_metrics} summarizes the 
total amount of time $T_{<S_p}$ the robot does not respect the separation distance $S_p$ (averaged over experiments, each lasting above one minute).

Increasing either $\sigma$ or $f_{\max}$ alone (profiles 1–3) does not substantially affect $d_{\min}^{RH}$, whereas increasing both (profile 4) yields a larger $d_{\min}^{RH}$. This effect is more pronounced for participants who moved cautiously, while for participants who moved faster the improvement was smaller, leading to larger variance in profile 4.
$T_{<S_p}$ decreases as $f_{\max}$ increases, and further when $\sigma$ increases, reaching a minimum when both parameters increased.
This highlights that SSM compliance can be influenced directly by control parameter settings.

Finally, we examined the effect of the damping mechanism described in Section \ref{sec:vm}.
Five experiments with different participants were performed using spring (profile 1) and damper ($c_0=150\,Nsm^{-1}, R=0.5\,m, f_{max}=50\,N$). 
The damper reacts proportionally to the human hand’s velocity, thereby providing stronger repulsion when the hand approaches the robot at high speed and reducing the robot’s motion toward the hand.
This reduced both $d_{\min}^{RH}$ and $T_{<S_p}$ compared to the spring-only baseline (profile 1.).
Responsiveness would likely be improved with higher-rate perception, currently limited at 10 Hz.

\begin{table}[h]
\centering
\caption{Minimum separation $d_{\min}^{RH}$ and violation duration $T_{<S_p}$ under different avoidance profiles and mechanisms.}
\label{tab:avoidance_metrics}

\begin{subtable}[t]{\columnwidth}
\label{tab:avoidance_metrics_1}
\centering
\begin{tabular}{lcc}
\hline
\textbf{Spring profile} & \textbf{$d_{\min}$ (cm)} & \textbf{$T_{<S_p}$ (s)} \\ 
\hline
Spring profile 1 (baseline) & $21.8\pm2.0$ & $4.6\pm1.8$ \\ 
Spring profile 2 (increased $f_{max}$) & $24.4\pm4.7$ & $2.3\pm1.1$ \\
Spring profile 3 (increased $\sigma$) & $24.8\pm9.1$ & $0.9\pm0.9$ \\
Spring profile 4 (both increased) & $30.0\pm9.8$ & $0.5\pm0.5$ \\
Spring profile 1 and damper & $19.9\pm2.9$ & $4.1\pm0.5$ \\ 
\hline
\end{tabular}
\vspace{-2mm}
\end{subtable}
\end{table}

\subsection{Conflict resolution} \label{sec:exp2}
While robots with non-conflicting tasks resolve interaction mechanically as shown in Section \ref{sec:exp1}, in general robots sharing a common workspace may stall. This typically occurs when two robots have incompatible tasks and symmetric policies to realize them.
We examine such cases and show how the prioritization strategy of Section \ref{stuck_detection} resolves them.

\begin{figure}[htbp]
    \centering
    \includegraphics[width=0.8\linewidth]{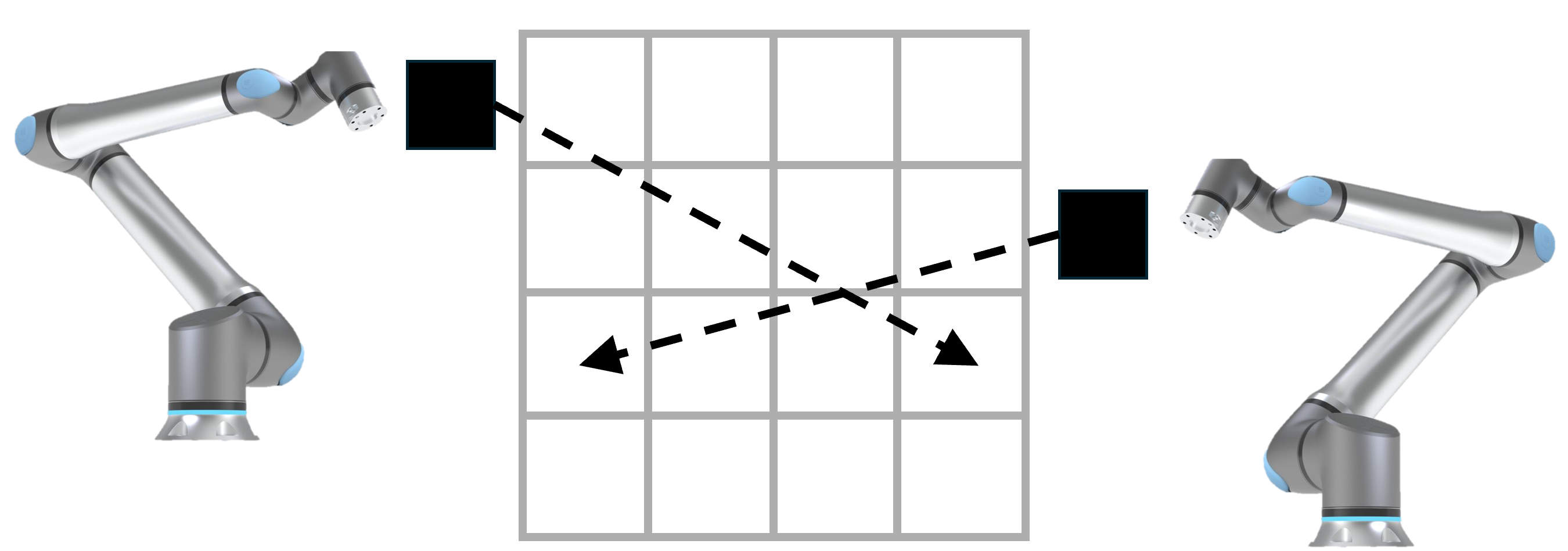}
    \caption{
    % Two robots transporting blocks along intersecting trajectories. This scenario induces a potential conflict.
    Two UR5 robots transport cubic blocks (black squares) toward their target cells (dashed arrows).
    The intersection of trajectories represents the potential conflict region.}
    \vspace{-2mm}
    \label{fig:conflic}
\end{figure}

\begin{figure*}[htbp]
    \centering
    \vspace{2mm}
    \begin{subfigure}[h]{.24\linewidth}
	\includegraphics[width=.99\columnwidth]{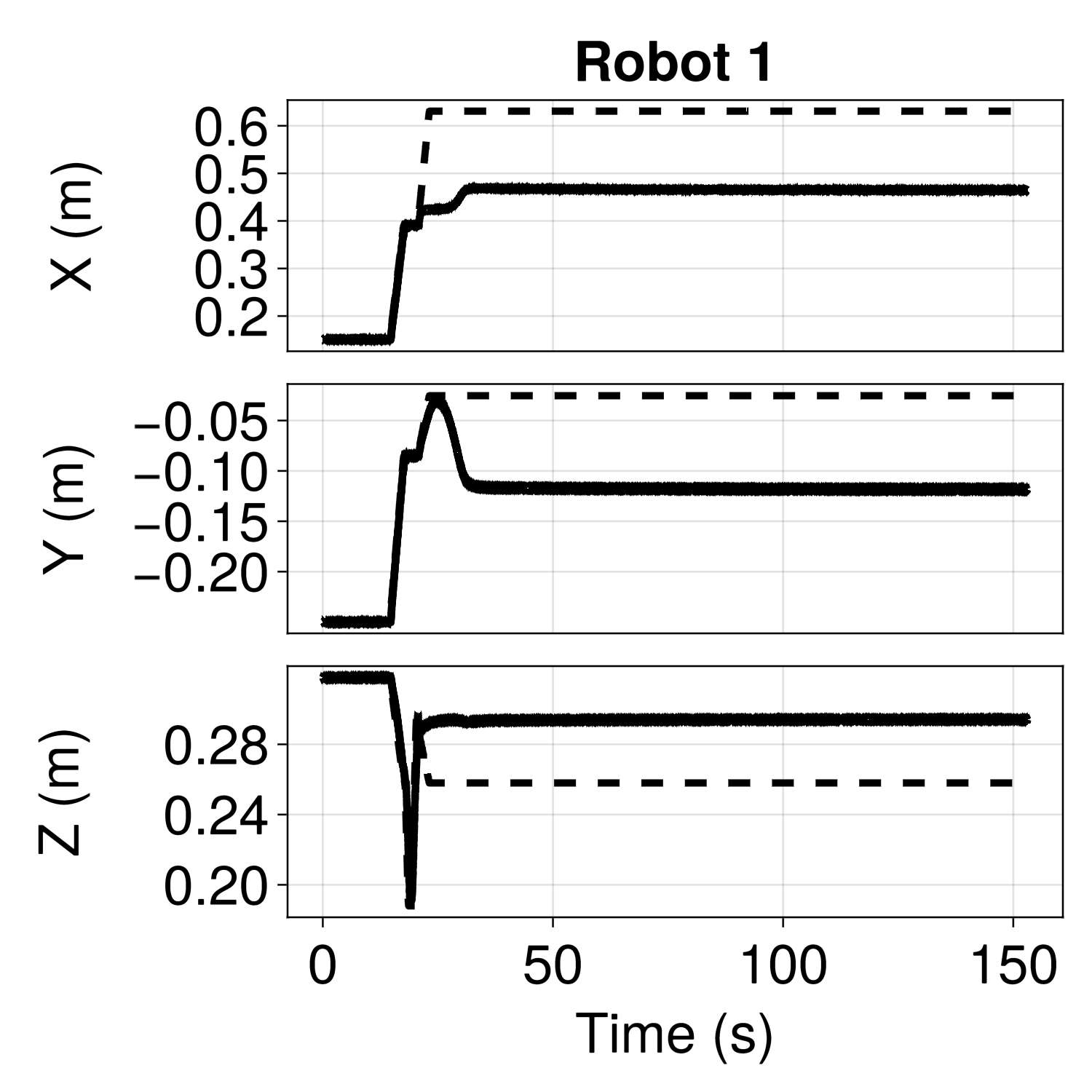}
    \end{subfigure}
    \begin{subfigure}[h]{.24\linewidth}
	\includegraphics[width=.99\columnwidth]{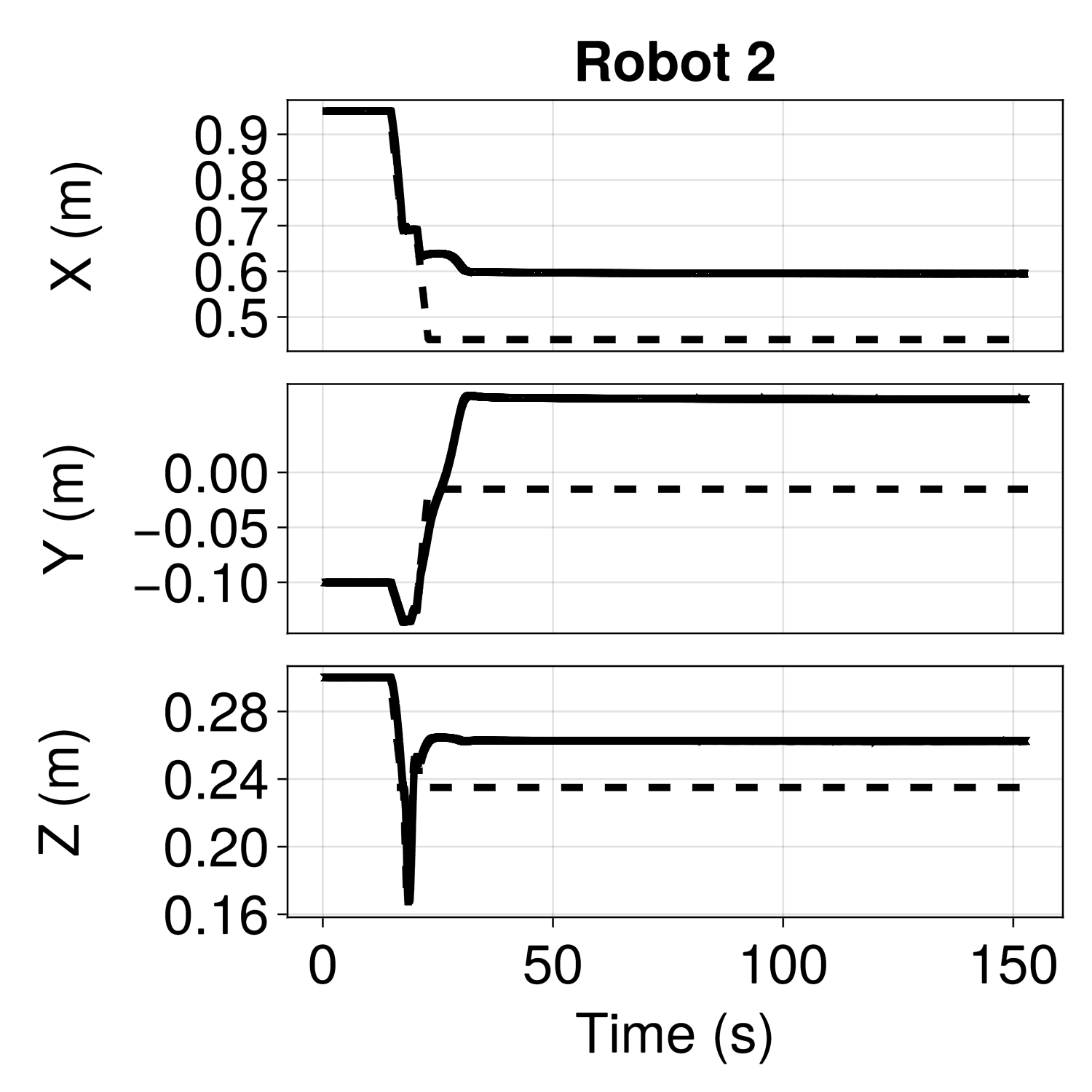}
    \end{subfigure}
    \begin{subfigure}[h]{.48\linewidth}
    \centering
	\includegraphics[width=.83\columnwidth]{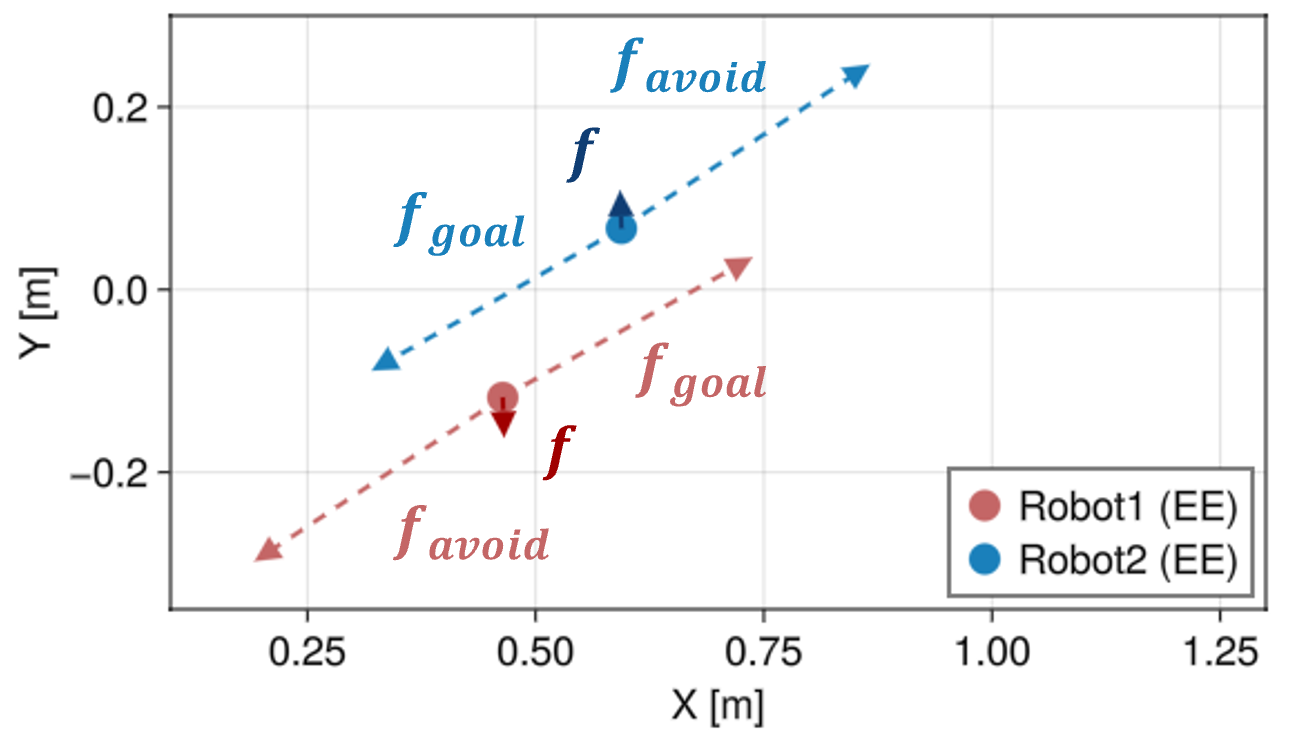}
    \end{subfigure}

    \begin{subfigure}[h]{.24\linewidth}
        \includegraphics[width=.99\columnwidth]{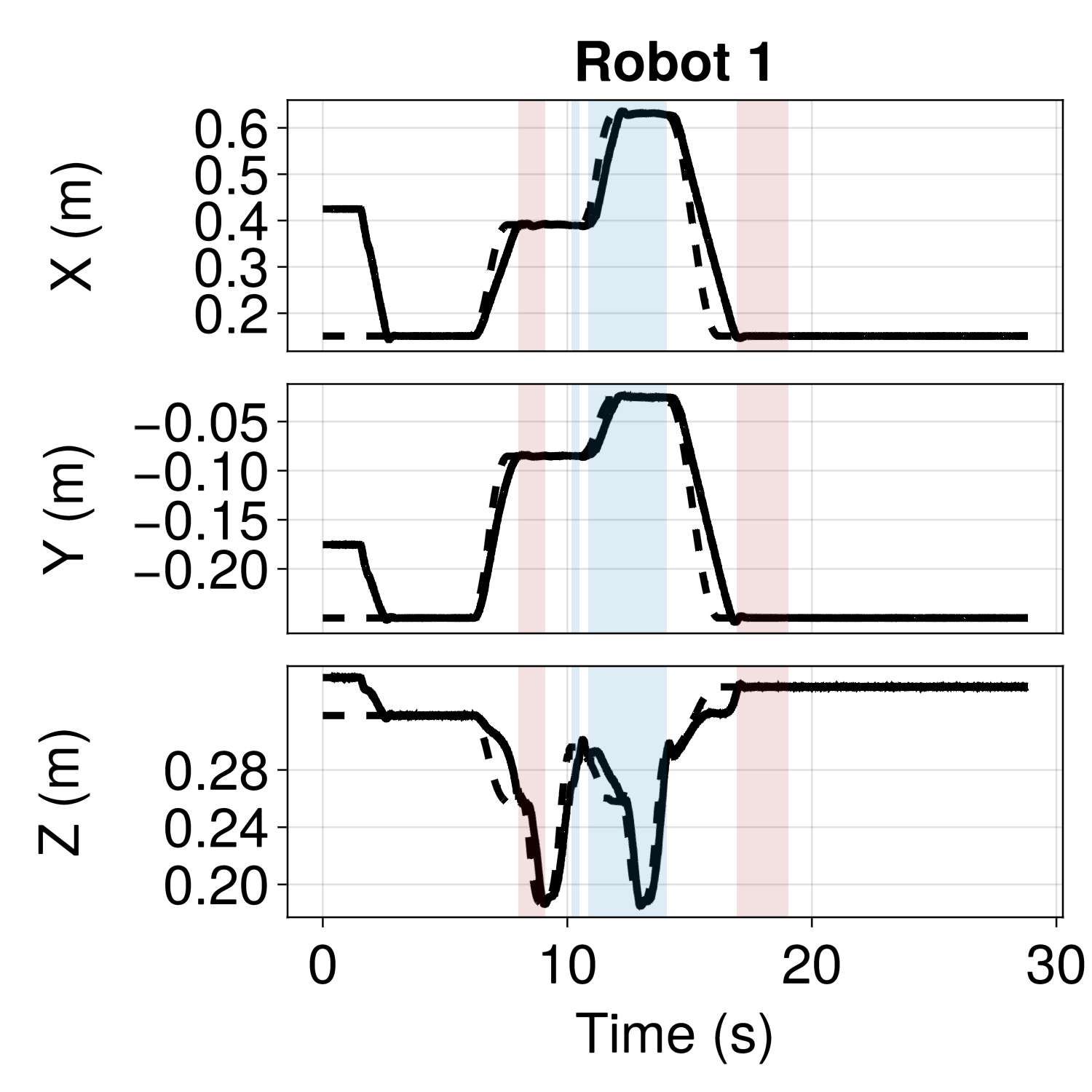}
    \end{subfigure}
    \begin{subfigure}[h]{.24\linewidth}
        \includegraphics[width=.99\columnwidth]{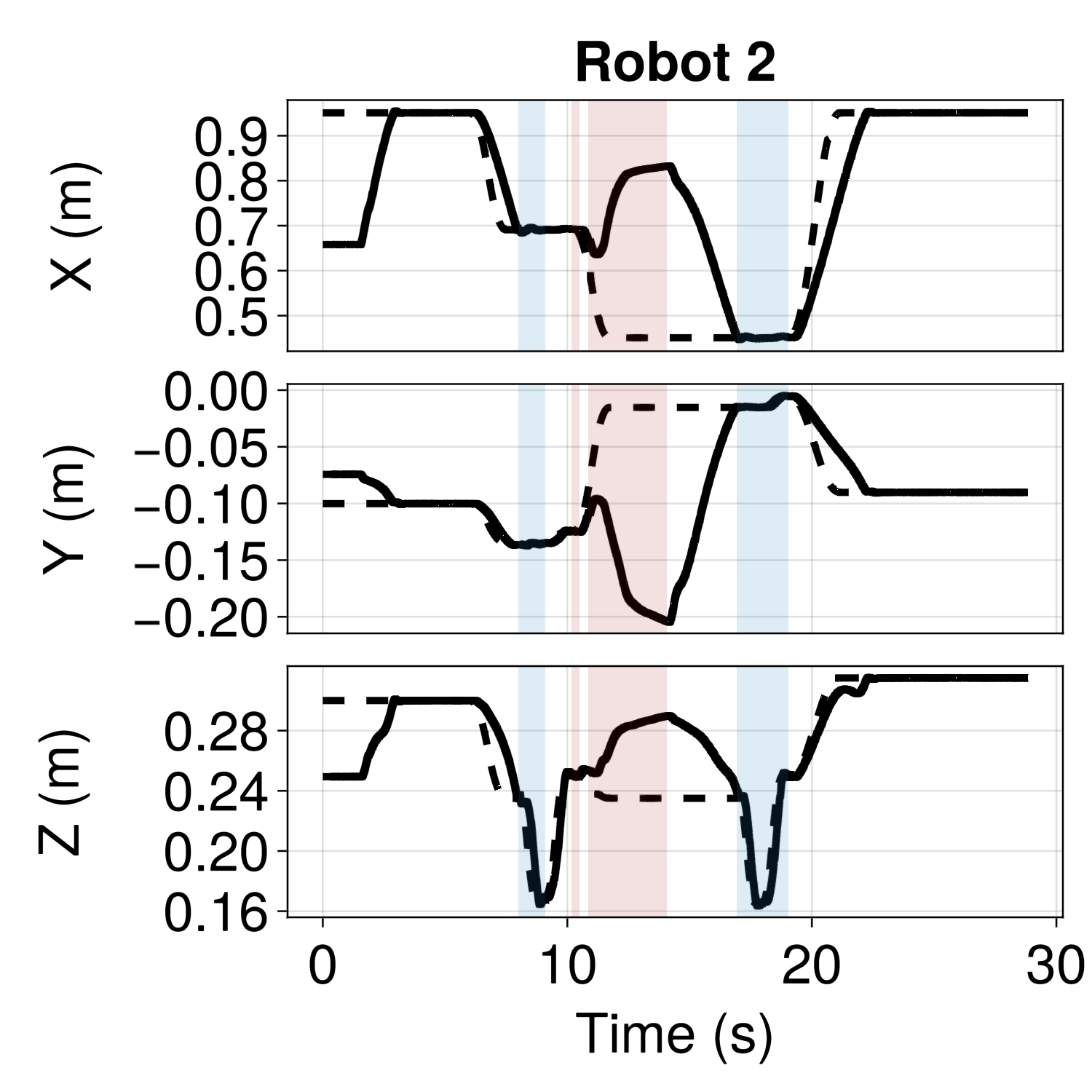}
    \end{subfigure}
    \begin{subfigure}[h]{.48\linewidth}
    \centering
	\includegraphics[width=.9\columnwidth]{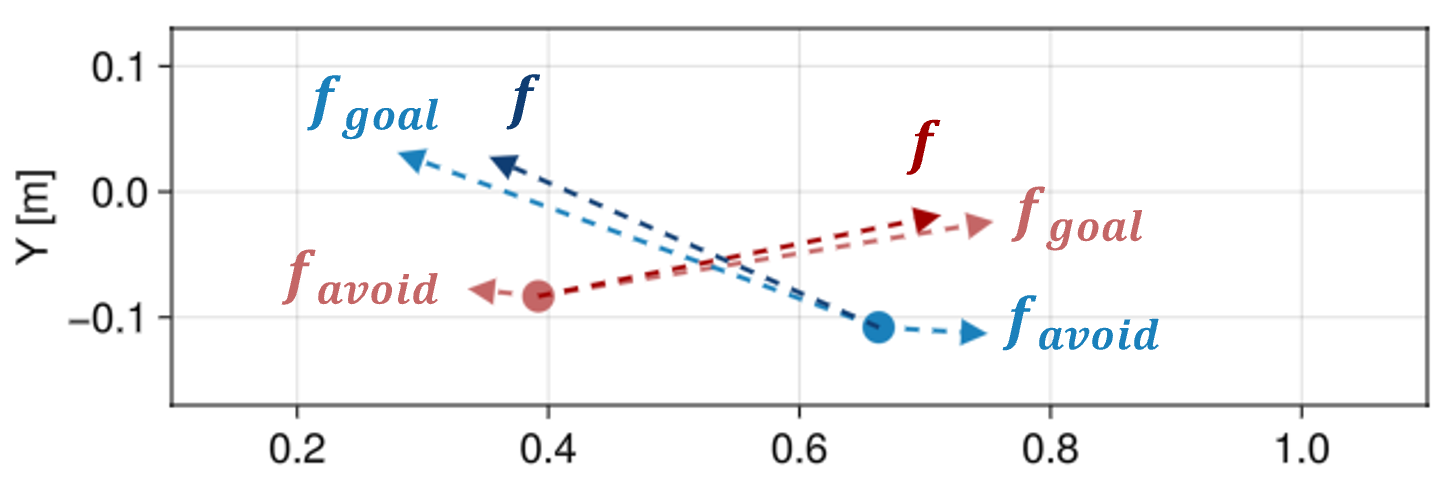}
	\includegraphics[width=.9\columnwidth]{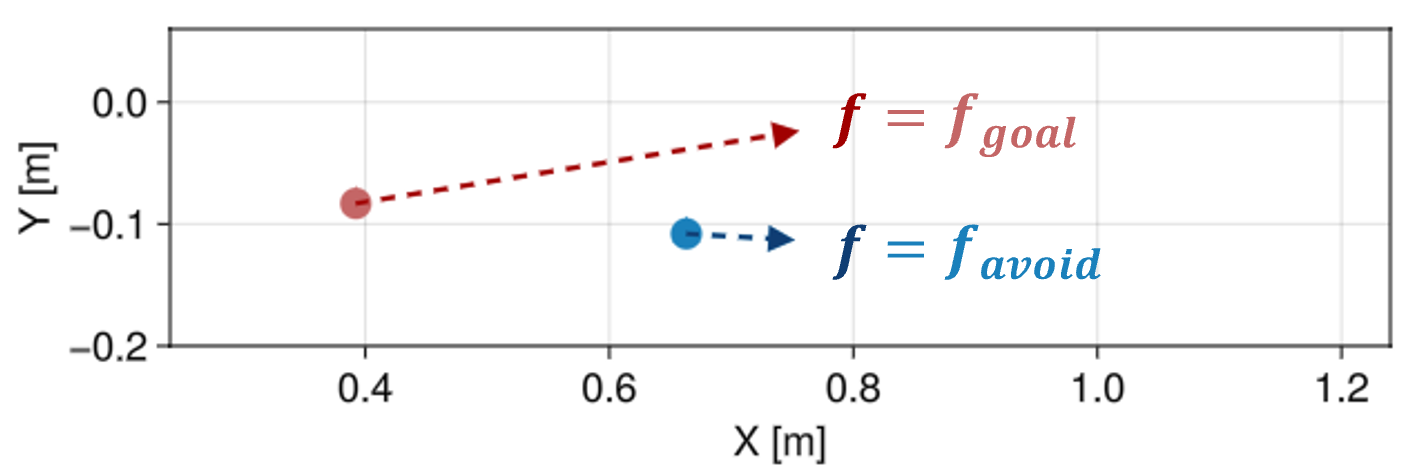}
    \end{subfigure}
    \vspace{-3mm}
    \caption{Top: without conflict resolution layer, robots entered a deadlock due to force balance at $t=25$ s and did not recover. Bottom: with the conflict resolution, robots resolved the conflict by prioritizing robot 1 at 10 s.
    Top-Left: robot trajectories (dashed - desired EE position; solid - EE position). 
    Top-Right: balancing forces at $t=150\,s$, indicating a stall; $\pmb{f}_{goal}$ goal-reaching force, $\pmb{f}_{avoid}$ cumulative avoidance force, $\pmb{f}=\pmb{f}_{goal}+\pmb{f}_{avoid}$. 
    Bottom-Left: robot trajectories.
    % (dashed - desired EE position; solid - EE position). 
    \textcolor{blue}{Blue} - prioritized interval, \textcolor{red}{Red} - non-prioritized interval. 
    Bottom-Right: top - stall detected at $t=10.87\,s$, bottom - stall resolved at $t=10.95\,s$}
    \label{fig:conflict_exp}
    \vspace{-4mm}
\end{figure*}

Consider the scenario in Fig. \ref{fig:conflic}, where two robots transport blocks along intersecting straight-line trajectories. 
Without negotiation, the attractive and avoidance forces balance near the intersection, resulting in a stall. This is illustrated through time-trajectories in Fig. \ref{fig:conflict_exp} (top-left). The motion of the robots stalls right after 25s, and the two-block task is still not completed after 150 s.
With our prioritization strategy, stalls are detected via the force-based metric \eqref{eq:rho} with the threshold for $\rho$ set to 4.0 N. A priority is negotiated, and spring stiffnesses are adjusted accordingly. This allows the robots to resolve the deadlock and complete the task successfully, as shown in  Fig. \ref{fig:conflict_exp} (bottom).

To analyze our conflict resolution strategy, we also consider a scenario in which two robots each take one of the 16 blocks and move it to one of the $4\times4$ inner grid along straight-line trajectories. A combinatorial enumeration with geometric intersection checks shows that, under random choices, the probability of two trajectories intersecting is 28.4\%. With three robots in the scene, the possibility increases up to 61.2 \% and with four robots it becomes even higher. Assuming that every intersection leads to a deadlock, conflicts would occur in nearly two thirds of tasks.
With the proposed stall detection and negotiation, we observed no unresolved deadlocks across all runs in the evaluated settings.

We further examined how control parameters influence task completion time, which is measured with elapsed time until the last block is placed. 
In this experiment, two robots placed all blocks into the grid as described in Section \ref{implementation}, without human involvement.
Three sets of `Slow', `Medium' and `Fast' parameters summarized in Table \ref{tab:exp2} yielded three distinct completion times, with five trials conducted for each case: $144.5\pm6.9$, $103.7\pm4.5$ and $88.5\pm4.8$ s.
Faster completion was achieved with stiffer springs connected to the goal with larger maximum forces and higher speeds in the moving-goal filter.

\subsection{Scalability} \label{sec:exp3}

\begin{figure*}[t]
    \centering
    \begin{subfigure}[h]{\linewidth}
    \centering
    \includegraphics[width=0.3\linewidth]{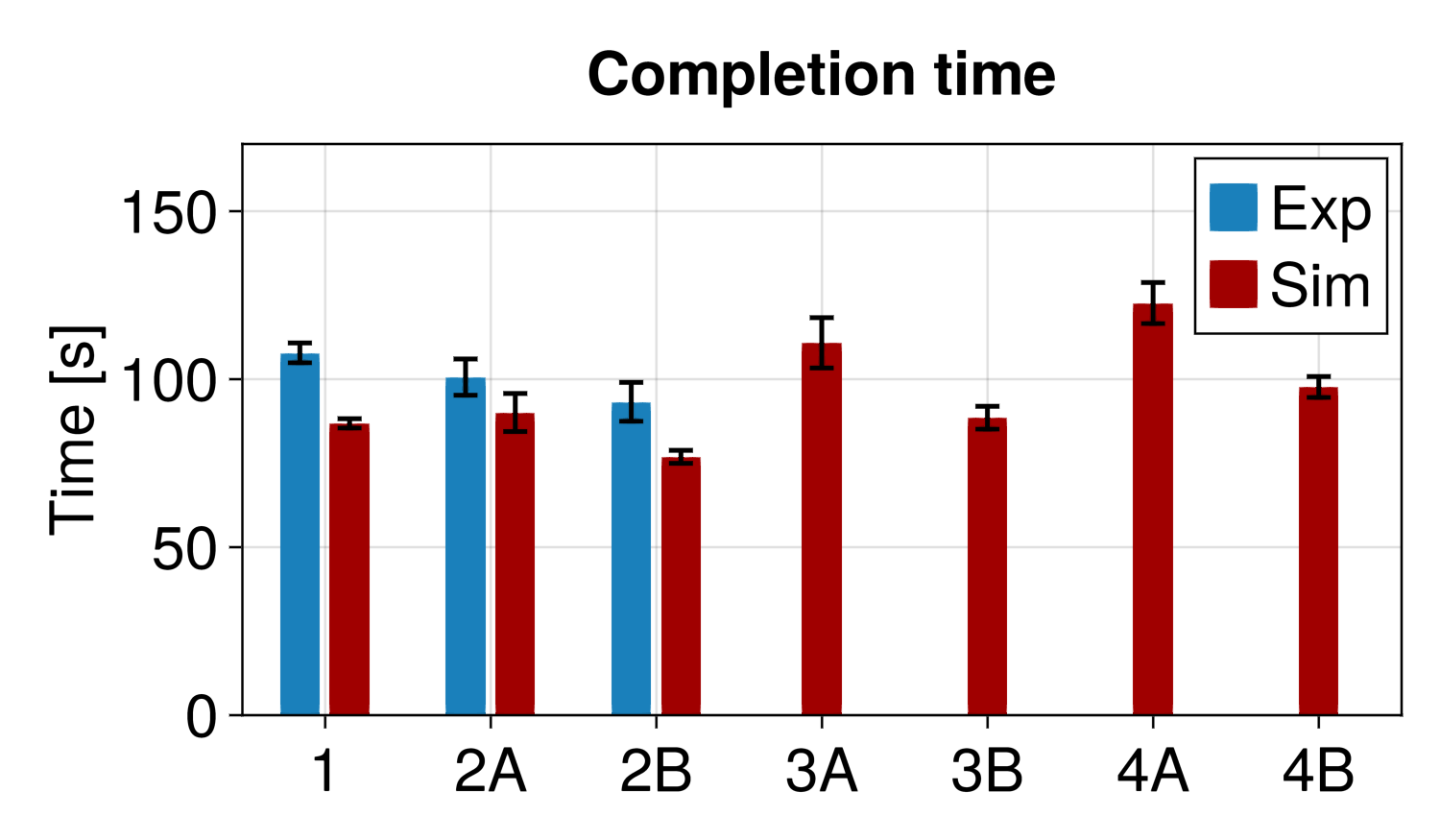}
    \includegraphics[width=0.61\linewidth]{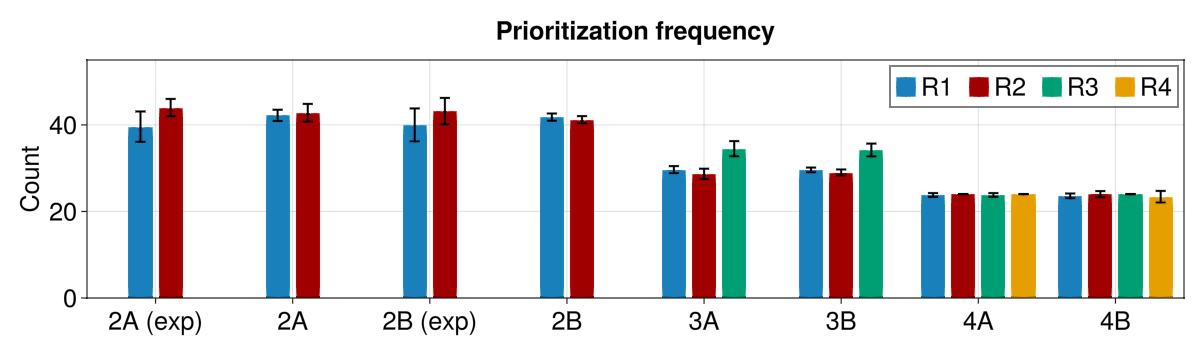} 
    \includegraphics[width=0.3\linewidth]{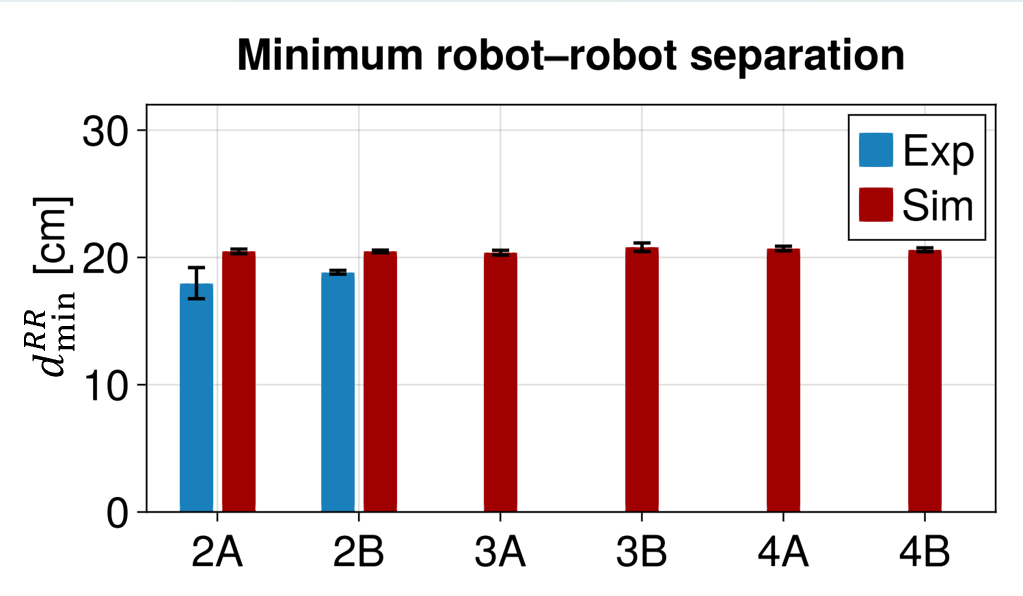}
    \includegraphics[width=0.61\linewidth]{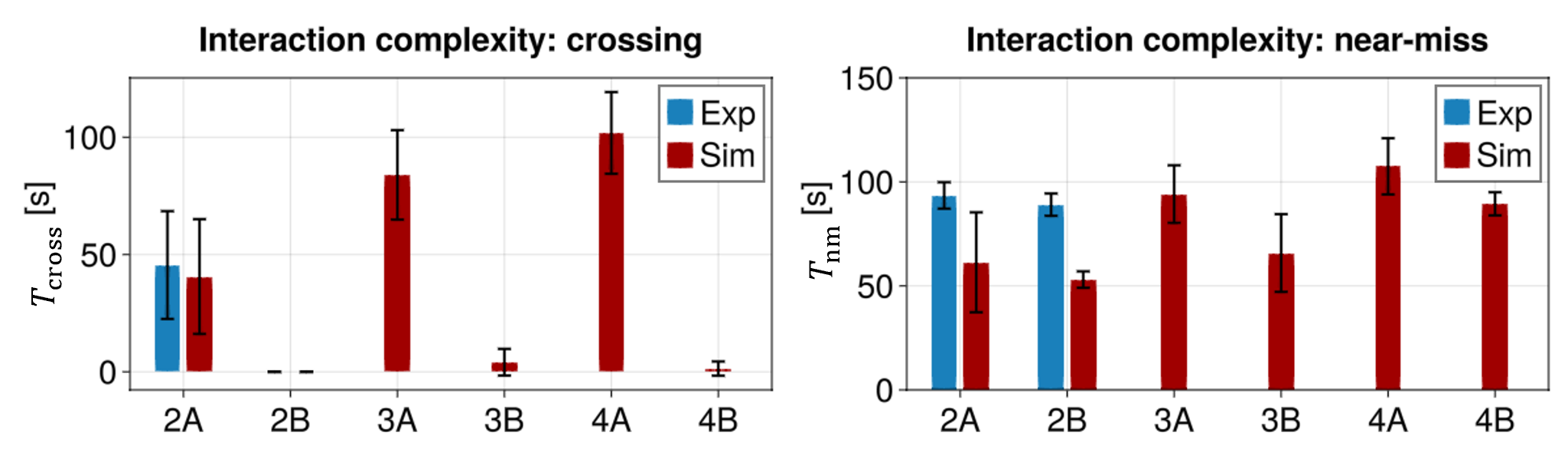} 
    \caption{\textbf{Multi-robot (real-world experiments and simulations).} Performance vs. team size. X-axis denotes number of robots and scenario. R$n$ denotes each robot in the simulation.}
    \label{fig:exp3-metric1}
    \end{subfigure}
    \begin{subfigure}[h]{\linewidth}
    \centering
    \includegraphics[width=0.65\linewidth]{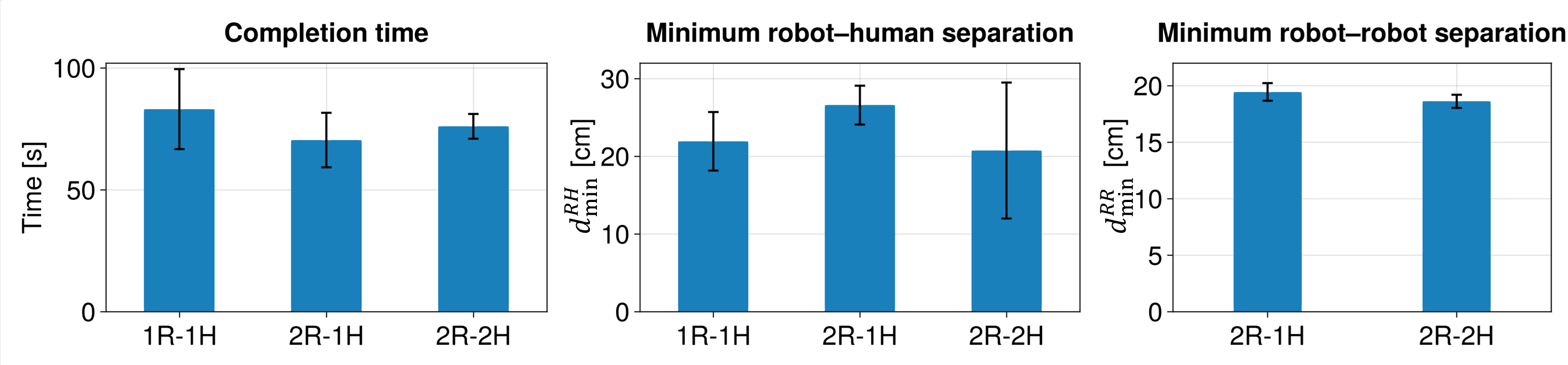} 
    \caption{\textbf{Multi-agent (robots and humans).} 
    % Completion time [s], minimum robot–human separation $d_{\min}^{RH}$ [cm], and robot–robot separation $d_{\min}^{RR}$ [cm] 
    Performance metrics vs. team size. $n$R-$m$H denotes collaboration between $n$ robots and $m$ humans.}
    \label{fig:exp3-metric2}
    \end{subfigure}
    \caption{
    \textbf{Completion time, safety and interaction metrics across scenarios and team sizes.} }
    \vspace{-6mm}
    \label{fig:metrics}
\end{figure*}

Finally, we assess the the ability of our virtual model control algorithm to handle robot team of different sizes. We consider two main scenarios.
\textbf{Scenario A}: all blocks are placed into grid according to Section \ref{implementation}.
\textbf{Scenario B} each robot places blocks only within its own half of the grid (reduced level of interaction). For each scenario and each team size, five experiments/simulations were conducted.

Simulations with three and four robots were conducted with the robots placed as shown in Fig. \ref{fig:four_robot_setup_sim}. 
In Scenario B with more than two robots, each robot was assigned a predefined range of blocks closer to its workspace.
Since real-world experiment time is influenced by hardware limitations, the completion time in real-world experiment is not directly comparable to completion time in simulation.
For consistency, we also report simulation results for the one- and two-robot cases.

To characterize interaction complexity, for every pair of simultaneous block transfers, we test if their ideal straight-line source-to-destination segments intersect (crossing event) or get very close (within $0.1$ m, near-miss event). We report the \emph{total crossing time} $T_\text{cross}$ and \emph{near-miss tim}e $T_\text{nm}$.
Prioritization frequency counts how often each robot is selected during negotiation, revealing potential asymmetries in allocation.

Results are summarized in Fig. \ref{fig:exp3-metric1}. 
$T_\text{cross}$ and $T_\text{nm}$ show that Scenario A involves more interaction between robots than Scenario B.
Both experiments and simulations show that the completion time improves up to two agents with reduced interaction (Scenario B); beyond this point, the limited workspace relative to robot size increases the completion time. Nevertheless, even in complex scenarios with four robots sharing the same workspace, all tasks were successfully completed without deadlocks. Prioritization counts were evenly distributed across robots, indicating fair workload allocation. Across all settings, robots successfully avoided collisions, maintaining an average separation of about 20 cm regardless of team size.

For mixed human–robot collaboration, the grid was partitioned into a checkerboard pattern: cells of one color were assigned to humans for block placement, while the remaining cells were reserved for robots. This ensures sufficient human–robot interactions. Scenario A is used when two robots were present.
For each team size, five experiments were conducted with different participants.

Task was successfully completed with different team size while maintaining around 20 cm between robots and robot-human as shown in Fig. \ref{fig:exp3-metric2}. 
Completion time improved slightly when increasing from two to three agents, but no further gains were observed with larger teams.

Our decentralized approach enabled seamless adaptation to different numbers of agents without altering the control structure: outside of conflict resolution, each robot relies only on the position of other agents which is obtainable with onboard camera. 
During the experiments, humans occasionally rested by withdrawing from the workspace and later re-entered to continue the task. The proposed framework naturally accommodates such changes, allowing collaboration despite dynamic team composition.

\section{Conclusions and Future Work} \label{conclusions}

This paper presented a decentralized, agent-agnostic, safety-aware framework for HRC based on VMC. Humans and robots share a common workspace. Motion planning and motion control are blended through virtual model control, enabling fast/slow reactive robot behavior based on intuitive parameter tuning. Deadlocks are detected through a force-balancing condition and resolved via a prioritization procedure that adapts the virtual model controller, reducing the probability of deadlocks. 
The method is highly scalable, with practical limits determined mainly by the computation of virtual forces and communication of sensor information.
The approach has been validated experimentally on position-controlled UR5 manipulators with up to two robots and two humans, and in simulation with up to four robots while maintaining inter-agent separation at around 20 cm, demonstrating (i) unified treatment of humans and robots, (ii) experimental deadlock-free operation, and (iii) scalability to larger teams.

In this paper, we focus on the distributed control architecture and therefore do not investigate how humans adapt their behavior in response (e.g., whether they become more assertive or more hesitant to enter the scene). 
A dedicated user study is an important direction for future work to evaluate how VMC shapes the collaborative workspace from the human perspective (usability, comfort, acceptance, etc.). 
Integration of human feedback within an adaptive virtual model control setting is also an important research direction. This could take into account non-verbal cues such as facial expressions or gestures to improve comfort, safety, and trust during human–robot interaction. The physical interpretability of VMC makes this adaptation tractable, offering a practical way toward user-centered collaboration without full redesign of the underlying controller.


%%%%%%%%%%%%%%%%%%%%%%%%%%%%%%%%%%%%%%%%%%%%%%%%%%%%%%%%%%%%%%%%%%%%%%%%%%%%%%%%

%%%%%%%%%%%%%%%%%%%%%%%%%%%%%%%%%%%%%%%%%%%%%%%%%%%%%%%%%%%%%%%%%%%%%%%%%%%%%%%%


\begin{table*}[htbp]
\centering
\vspace{3mm}
\caption{
Control parameters for each experiment.
}
\label{tab:control_parameters}
\begin{subtable}[t]{\textwidth}
\centering
\caption{Agent-agnostic and safety-aware collaboration (Section \ref{sec:exp1} and \ref{sec:exp1-2}).}
\label{tab:exp1}
\begin{tabular}{@{}cc|cc|cc|cc@{}}
    \multicolumn{6}{c|}{\textbf{Robot 1}} &
      \multicolumn{2}{c}{\textbf{Robot 2}}\\\hline
    \multicolumn{2}{c|}{\textbf{Goal spring}} &
      \multicolumn{2}{c|}{\textbf{Avoidance spring (for other robots)}} &
      \multicolumn{2}{c|}{\textbf{Avoidance spring (for human hand)}} &
      \multicolumn{2}{c}{\textbf{Goal spring}}\\
    Stiffness [N/m] & $f_{max}$ [N] &
    Stiffness [N/m] & $f_{max}$ [N] &
    $\sigma$ [m] & $f_{max}$ [N]  &
    Stiffness [N/m] & $f_{max}$ [N]\\ \hline
    1000 & 10 &
    -1000 & -40 & % sigma=0.066
    0.18 & -60 &
    3000 & 20 \\
\end{tabular}
\end{subtable}

\vspace{2mm}
\begin{subtable}[t]{\textwidth}
\centering
\caption{Conflict resolution (Section \ref{sec:exp2}). All robots have the same spring profiles. `Medium' is used for showcase of conflict resolution.}
\label{tab:exp2}
\begin{tabular}{@{}c|cc|c|cc@{}}
    & \multicolumn{2}{c|}{\textbf{Goal spring}} & \multicolumn{1}{c|}{\textbf{Filter goal}} &
      \multicolumn{2}{c}{\textbf{Avoidance spring (for other robots)}} \\
     &
    Stiffness [N/m] & $f_{max}$ [N] & Speed [m/s] &
    Stiffness [N/m] & $f_{max}$ [N]   \\ \hline
    Slow &
    2000 & 15 & 0.1 &
    -900 & -40 \\ \hline % sigma=0.066
    Medium &
    2000 & 20 & 0.4 &
    -900 & -40 \\ \hline
    Fast &
    3000 & 30 & 1.0 &
    -900 & -40 
\end{tabular}
\end{subtable}

\vspace{3mm}
\begin{subtable}[t]{\textwidth}
\centering
\caption{Scalability (Section \ref{sec:exp3}). All robots have the same spring profiles.}
\label{tab:exp3}
\begin{tabular}{@{}c|cc|c|cc|cc@{}}
    & \multicolumn{2}{c|}{\textbf{Goal spring}} & \multicolumn{1}{c|}{\textbf{Filter Goal}} &
      \multicolumn{2}{c|}{\textbf{Avoidance spring (for other robots)}} &
      \multicolumn{2}{c}{\textbf{Avoidance spring (for human hand)}} \\
    \textbf{Exp} &
    Stiffness [N/m] & $f_{max}$ [N] &
    Speed [m/s] &
    Stiffness [N/m] & $f_{max}$ [N] &
    $\sigma$ [m] & $f_{max}$ [N]  \\ \hline
    3 &
    2000 & 20 &
    0.4&
    -900 & -40 & % sigma=0.066
    0.18 & -60 \\
\end{tabular}
\end{subtable}
\vspace{-5mm}
\end{table*}

%%%%%%%%%%%%%%%%%%%%%%%%%%%%%%%%%%%%%%%%%%%%%%%%%%%%%%%%%%%%%%%%%%%%%%%%%%%%%%%%

\bibliographystyle{IEEEtran}
\bibliography{IEEEexample}

\end{document}